\documentclass[10pt,twocolumn,letterpaper]{article}

\usepackage{cvpr}              

\usepackage{multirow}
\usepackage{booktabs}
\usepackage[table]{xcolor}
\usepackage{graphicx}
\usepackage{svg}
\usepackage{tabularx}

\definecolor{cvprblue}{rgb}{0.21,0.49,0.74}
\usepackage[pagebackref,breaklinks,colorlinks,allcolors=cvprblue]{hyperref}

\def\paperID{*****} 
\def\confName{CVPR}
\def\confYear{2026}

\title{Breaking Spurious Correlations via Generative Randomization \\
and Cross-Variant Self-Supervised Learning}

\author{
Suraj Yadav \quad
Anjaneya Sharma\thanks{Equal contribution.} \quad
Siddharth Yadav\footnotemark[1] \\
Indraprastha Institute of Information Technology Delhi \\
{\tt\small \{suraj24098, anjaneya21449, siddharth23525\}@iiitd.ac.in}
}

\begin{document}
\maketitle
\begin{abstract}
Deep neural networks trained with Empirical Risk Minimization (ERM) often fail under distribution shifts because they exploit spurious correlations between object labels and background context. Recent generative approaches address this issue by creating counterfactual images with altered contexts, but typically use these samples as standard data augmentation, leaving the model free to retain background-sensitive representations. We propose a two-stage framework that uses generative intervention to explicitly learn background-invariant visual representations. First, we isolate the foreground object using zero-shot segmentation and generate context-shifted variants with a structure-preserving diffusion model, preserving object identity while varying the surrounding environment. We then introduce Cross-Variant Self-Supervised Learning, where variants of the same object under different backgrounds form positive pairs in a contrastive objective. This encourages the encoder to align object-centric representations while suppressing background-specific cues. Then, we fine-tune the pretrained encoder using an ERM warm-up followed by GroupDRO with layer-wise learning rates. Experiments on distribution-shift benchmarks demonstrate best worst-group performance, achieving 92.5\% on Waterbirds, 81.7\% on MetaShift, and 87.4\% on NICO++. \href{https://github.com/surajyadav-research/GRSSL}{GitHub}
\end{abstract}

\section{Introduction}
\label{sec:intro}

Deep neural networks trained with Empirical Risk Minimization (ERM) achieve strong performance when training and test data are drawn from similar distributions, but often degrade under distribution shift \cite{arjovsky2020invariantriskminimization, hendrycks2021naturaladversarialexamples}. A major cause of this failure is the reliance on spurious correlations: predictive but non-causal associations between labels and visual attributes such as background, texture, or scene context \cite{geirhos2022imagenettrainedcnnsbiasedtexture}. For instance, in the Waterbirds benchmark, waterbirds are frequently associated with water backgrounds and landbirds with land backgrounds. An ERM-trained classifier can therefore learn to rely on the background rather than bird-specific features, leading to poor performance when the same bird categories appear in atypical contexts \cite{beery2018recognitionterraincognita, sagawa2020distributionallyrobustneuralnetworks}.

Prior work addresses this problem through either training-time reweighting or data augmentation. Methods such as GroupDRO \cite{sagawa2020distributionallyrobustneuralnetworks} and JTT \cite{liu2021justtraintwiceimproving} improve robustness by emphasizing difficult or underperforming groups. However, these methods can be sensitive to the quality of group supervision and to early optimization dynamics, especially when minority groups are small or noisy. More recent generative approaches synthesize counterfactual samples to reduce group imbalance and weaken spurious correlations \cite{parast2025ddbdiffusiondrivenbalancing}. While effective, these methods typically treat generated images as additional supervised training samples. This can improve dataset coverage, but it does not explicitly force the encoder to represent the same object consistently across different contexts.

We argue that counterfactual generation is most useful when paired with an objective that directly learns invariance across generated context shifts. To this end, we propose a two-stage framework that combines generative intervention with representation-level invariance learning. In the first stage, we isolate the foreground causal object using a zero-shot segmentation pipeline based on Grounding DINO \cite{liu2024groundingdinomarryingdino} and Segment Anything Model (SAM) \cite{kirillov2023segment}. We then use a structure-preserving diffusion inpainting model to generate context-shifted variants by modifying the background while preserving the object identity.

\begin{figure*}[t]
  \centering
  \includegraphics[width=1\linewidth]{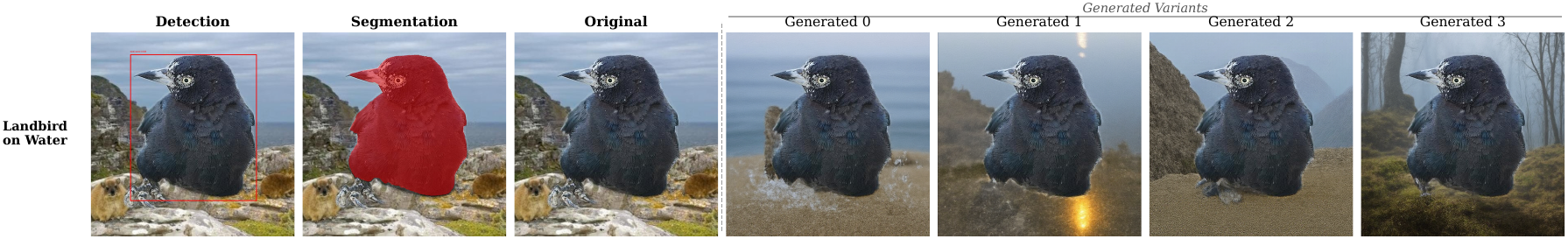}
  \caption{Examples of generated variants produced by replacing the background while preserving the foreground object identity.}
  \label{fig:detail}
\end{figure*}

Rather than using these generated images only as independent augmentations, we introduce Cross-Variant Self-Supervised Learning. For each image, generated variants that share the same foreground object but differ in background context are treated as positive pairs in a contrastive objective. This encourages the encoder to retain object-relevant information while suppressing background-specific cues. Then, we fine-tune the SSL-pretrained encoder with an ERM warm-up followed by GroupDRO. We use layer-wise learning rates to preserve invariant features learned during pretraining while adapting the classifier to the target task.

Our contributions are summarized as follows:

\begin{itemize}
    \item We introduce a high-fidelity generative intervention pipeline that isolates foreground causal objects and produces diverse context-shifted variants using structure-preserving diffusion.

    \item We propose Cross-Variant Self-Supervised Learning, a contrastive pretraining strategy that aligns representations of the same object across generated backgrounds to reduce background reliance, followed by ERM warm-up, layer-wise fine-tuning, and GroupDRO for robust optimization.

    \item We achieve strong worst-group performance across multiple distribution-shift benchmarks, reporting 92.5\% on Waterbirds, 81.7\% on MetaShift, and 87.4\% on NICO++, while maintaining high average accuracies of 95.4\%, 82.6\%, and 94.0\%, respectively.
\end{itemize}
\section{Methodology}
\label{sec:method}

We formulate our approach as a two-stage framework for mitigating spurious correlations. In Stage 1, we isolate the foreground object and generate context-shifted variants by randomizing the surrounding background. In Stage 2, we use these variants for Cross-Variant Self-Supervised Learning to encourage background-invariant representations, followed by ERM-warmup GroupDRO fine-tuning with layer-wise learning rates to improve worst-group generalization.

\subsection{Problem Formulation}
Consider a training dataset $\mathcal{D} = \{(x_i, y_i, e_i)\}_{i=1}^N$, where $x_i \in \mathcal{X}$ is an image, $y_i \in \mathcal{Y}$ is the target label, and $e_i$ denotes the environment or context attribute. We define the class--environment group as $g_i = (y_i, e_i)$, where $g_i \in \mathcal{G}$, which is used for worst-group evaluation and GroupDRO optimization. From a Structural Causal Model (SCM) \cite{Pearl_2009} perspective, each image $x_i$ is composed of causal features $c_i$ (e.g., the bird) and spurious features $s_i$ (e.g., the background). Standard ERM minimizes the average loss $\frac{1}{N}\sum_{i=1}^N \ell(f(x_i), y_i)$, which inadvertently allows the model $f$ to exploit dataset-specific correlations where $P(y|s) \neq P(y)$. Our objective is to learn a representation that relies on $c_i$, maximizing the worst-group accuracy over all class--environment groups $g \in \mathcal{G}$.

\subsection{Generative Environmental Randomization}
To decouple the object $c_i$ from its spurious environment $s_i$, we employ a zero-shot detection and segmentation pipeline. For a given image $x_i$ and class label $y_i$, we query Grounding DINO \cite{liu2024groundingdinomarryingdino} to extract a bounding box $B_i$. We then pass $B_i$ to the Segment Anything Model (SAM) \cite{kirillov2023segment} to obtain a pixel-level binary mask $M_i$ isolating the causal object.

Instead of merely swapping backgrounds between existing dataset classes, we inject semantic diversity. We invert the mask to target the background region, $\bar{M}_i$, and utilize FLUX.1-Fill \cite{labs2025flux1kontextflowmatching, flux2024}, a state-of-the-art structural diffusion model, to generate a set of context-shifted variants $V_i = \{x_i^{(1)}, x_i^{(2)}, \dots, x_i^{(K)}\}$. Each variant $x_i^{(k)}$ is conditioned on a distinct textual prompt $p_k$ (e.g., ``lake'', ``mountain'', ``forest''). The forward diffusion and denoising process alters only the masked region, preserving the pixel structure of $c_i$ while completely resynthesizing $s_i$.

\begin{figure*}[t]
  \centering
  \includegraphics[width=0.95\linewidth]{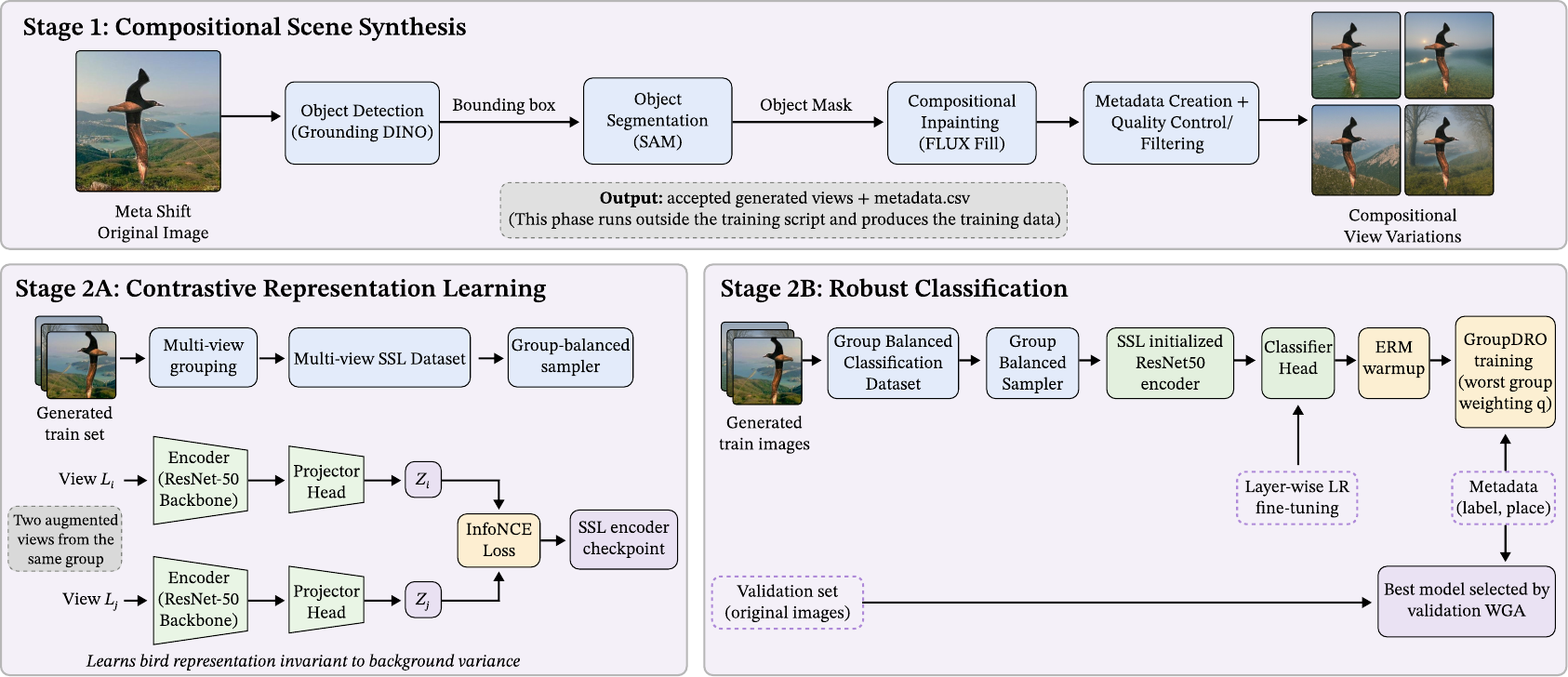}
  \caption{Stage 1 generates background-randomized variants using object detection, segmentation, and diffusion inpainting. Stage 2 learns background-invariant representations through Cross-Variant SSL, then fine-tunes the encoder using ERM warmup followed by GroupDRO.}
  \label{fig:pipeline}
\end{figure*}

\subsection{Cross-Variant Self-Supervised Learning}
\label{subsec:ssl}

Standard contrastive learning paradigms, such as SimCLR \cite{chen2020simpleframeworkcontrastivelearning}, generate positive pairs $(z_1, z_2)$ by applying stochastic data augmentations (e.g., color jitter, cropping) to the \textit{same} image. While effective for low-level invariances, this does not explicitly enforce semantic background invariance.

We introduce Cross-Variant SSL. Let $E_\theta$ be our encoder and $P_\phi$ be an MLP projector. During training, for a given causal object $i$, we sample two distinct generative variants $x_i^{(u)}, x_i^{(v)} \in V_i$, where $u \neq v$. After applying independent light augmentations $t_u,t_v \sim \mathcal{T}$, we compute the projections
\[
z_i^{(u)} = P_\phi(E_\theta(t_u(x_i^{(u)}))),
\qquad
z_i^{(v)} = P_\phi(E_\theta(t_v(x_i^{(v)}))).
\]

Since the two views share the same foreground object but contain different backgrounds, maximizing their similarity encourages the encoder to focus on object-relevant features and reduce reliance on background-specific cues. We optimize the normalized temperature-scaled cross-entropy (NT-Xent) \cite{chen2020simpleframeworkcontrastivelearning} loss:
\begin{equation}
\mathcal{L}_{\mathrm{SSL}}
=
\frac{1}{2B}
\sum_{m=1}^{2B}
-\log
\frac{
\exp(\mathrm{sim}(z_m, z_{p(m)})/\tau)
}{
\sum_{n=1}^{2B}
\mathbb{1}_{[n \neq m]}
\exp(\mathrm{sim}(z_m,z_n)/\tau)
}
\end{equation}
where $B$ is the batch size, $\mathrm{sim}(\cdot, \cdot)$ is cosine similarity, $\tau$ is the temperature scalar, and $(z_a,z_b)$ denotes a positive pair sampled from two variants of the same causal object.

\subsection{Fine-Tuning: ERM-Warmup GroupDRO}
We discard the SSL projection head and initialise a linear classification head $H_\psi$. To map the background-invariant features to class labels without destroying the representations learned in SSL, we employ layer-wise learning rates.

To prioritize worst-group performance, we optimize the network using Group Distributionally Robust Optimization (GroupDRO) \cite{sagawa2020distributionallyrobustneuralnetworks}. GroupDRO minimizes the loss over the worst-case group $g \in \mathcal{G}$:
\begin{equation}
\mathcal{L}_{\text{DRO}}(\theta, \psi) = \max_{q \in \Delta_{|\mathcal{G}|}} \sum_{g \in \mathcal{G}} q_g \, \mathbb{E}_{(x,y) \sim \mathcal{D}_g} \left[\ell(H_\psi(E_\theta(x)), y)\right]
\end{equation}
where $q$ is an adversarially updated group weight vector parameterized by a step size $\eta_{q}$.

Because GroupDRO is highly sensitive to random initialization and noisy minority samples early in training, we introduce an \textit{ERM Warmup} phase. For the first $W$ epochs, we fix $q_g = \frac{1}{|\mathcal{G}|}$, which corresponds to group-balanced ERM under our balanced sampler. Once the classifier head $H_\psi$ has stabilized, we unlock the multiplicative weights update:
\begin{equation}
q_g \leftarrow \frac{q_g \exp(\eta_q \mathcal{L}_g)}{\sum_{g' \in \mathcal{G}} q_{g'} \exp(\eta_q \mathcal{L}_{g'})}
\end{equation}
seamlessly transitioning into strict worst-group optimization for the remainder of the training phase.
\section{Experiments}
\label{sec:experiments}

We evaluate our proposed framework on standard vision benchmarks designed to assess spurious correlations and out-of-distribution (OOD) generalization. We compare our performance against relevant baselines, both with and without data augmentation, and conduct extensive ablation studies to isolate the impact of our proposed components.
\begin{table*}[t]
\centering

\normalsize

\setlength{\tabcolsep}{6pt}

\renewcommand{\arraystretch}{1.2}

\caption{Comparison of worst-group accuracy and average accuracy across three standard spurious correlation datasets. The ``Group Info'' column indicates whether group labels are used during training and validation. We report the mean and standard deviation over three runs. Best results are highlighted in \textbf{bold}.}
\label{tab:main_results}

\begin{tabular}{lccccccc}
\toprule
\multirow{2}{*}{\textbf{Method}} 
& \multirow{2}{*}{\textbf{train/val}} 
& \multicolumn{2}{c}{\textbf{Waterbirds}} 
& \multicolumn{2}{c}{\textbf{MetaShift}} 
& \multicolumn{2}{c}{\textbf{NICO++}} \\
\cmidrule(lr){3-4} \cmidrule(lr){5-6} \cmidrule(lr){7-8}
&  & \textbf{Worst} & \textbf{Average} 
& \textbf{Worst} & \textbf{Average} 
& \textbf{Worst} & \textbf{Average} \\
\midrule

Group DRO \cite{sagawa2020distributionallyrobustneuralnetworks} 
& $\checkmark/\checkmark$ 
& $91.6_{\pm1.3}$ 
& $93.4_{\pm0.4}$ 
& $67.1_{\pm3.5}$ 
& $72.9_{\pm2.3}$ 
& $76.1_{\pm1.5}$ 
& $91.3_{\pm0.7}$ \\

DFR
\cite{kirichenko2023layerretrainingsufficientrobustness} 
& $\times/\checkmark$ 
& $92.3_{\pm0.4}$ 
& $93.3_{\pm0.5}$ 
& $72.8_{\pm0.6}$ 
& $77.5_{\pm0.6}$ 
& $87.3_{\pm3.9}$ 
& $93.7_{\pm1.0}$ \\

\midrule

LISA \cite{yao2022improvingoutofdistributionrobustnessselective} 
& $\checkmark/\checkmark$ 
& $89.2_{\pm0.6}$ 
& $91.8_{\pm0.3}$ 
& $59.8_{\pm2.3}$ 
& $70.0_{\pm0.7}$ 
& $81.9_{\pm2.2}$ 
& $90.2_{\pm2.3}$ \\

DaC \cite{noohdani2024decomposeandcomposecompositionalapproachmitigating} 
& $\times/\checkmark$ 
& $92.4_{\pm0.4}$ 
& $95.3_{\pm0.4}$ 
& $78.3_{\pm1.6}$ 
& $79.6_{\pm0.1}$ 
& $84.9_{\pm3.2}$ 
& $93.7_{\pm1.0}$ \\

DDB \cite{parast2025ddbdiffusiondrivenbalancing} 
& $\times/\checkmark$ 
& $88.37_{\pm1.8}$ 
& $94.2_{\pm0.5}$ 
& $78.2_{\pm2.2}$ 
& $79.3_{\pm0.1}$ 
& $80.9_{\pm4.6}$ 
& $92.2_{\pm0.7}$ \\

\midrule

ERM 
& $\times/\times$ 
& $70.8_{\pm0.5}$ 
& $91.6_{\pm0.1}$ 
& $61.3_{\pm3.4}$ 
& $73.9_{\pm1.5}$ 
& $74.9_{\pm1.8}$ 
& $90.9_{\pm0.3}$ \\

\textbf{GRSSL} 
& $\checkmark/\checkmark$ 
& $\mathbf{92.5_{\pm0.3}}$ 
& $\mathbf{95.4_{\pm0.4}}$ 
& $\mathbf{81.7_{\pm2.2}}$ 
& $\mathbf{82.6_{\pm1.2}}$ 
& $\mathbf{87.4_{\pm3.8}}$ 
& $\mathbf{94.0_{\pm0.5}}$ \\

\bottomrule
\end{tabular}
\end{table*}

\subsection{Evaluation Settings}
\textbf{Datasets.} We evaluate our method on three challenging benchmarks characterized by severe environmental distribution shifts: \textbf{Waterbirds} \cite{sagawa2020distributionallyrobustneuralnetworks}, \textbf{MetaShift} \cite{liang2022metashiftdatasetdatasetsevaluating}, and \textbf{NICO++} \cite{zhang2022nicobetterbenchmarkingdomain}. Extended dataset details and specific subset definitions are provided in the Appendix~\ref{sec:dataset_details}.

\textbf{Implementation Details.}
We use an ImageNet-pretrained ResNet-50 \cite{he2015deepresiduallearningimage} backbone for all experiments. Causal objects are localized with Grounding DINO \cite{liu2024groundingdinomarryingdino} using a box threshold of $0.25$ and segmented with SAM \cite{kirillov2023segment}. FLUX.1-Fill-dev \cite{labs2025flux1kontextflowmatching, flux2024} generates four background-randomized variants per training image at $1024 \times 1024$ resolution; validation and test images are kept unchanged.

For each generated variant, the class label is inherited from the source image. The environment/context label is assigned according to the generation prompt category, and the GroupDRO group is defined as $g=(y,e)$. We train the encoder with Cross-Variant SSL for 10 epochs using an InfoNCE loss with temperature $\tau=0.2$ and a 128-dimensional projection head. In Stage 2, we remove the projection head and fine-tune for 6 epochs using a 2-epoch ERM warmup followed by GroupDRO with step size $\eta_q=0.01$. Additional implementation and dataset settings are provided in Appendix~\ref{sec:implementation_details}.

\subsection{Main Results}
Table~\ref{tab:main_results} reports worst-group and average accuracy on Waterbirds, MetaShift, and NICO++. Our method achieves strong performance across all benchmarks, improving worst-group robustness while maintaining high average accuracy. Unlike prior augmentation-based methods that use generated samples mainly as additional supervised data, our framework uses generated context variants to learn background-invariant representations through Cross-Variant SSL.

On Waterbirds, our method achieves $92.5 \pm 0.3\%$ worst-group accuracy and the highest average accuracy of $95.4 \pm 0.4\%$, improving over ERM by $21.7$ percentage points in worst-group accuracy. On MetaShift, it obtains the strongest worst-group and average accuracies, reaching $81.7 \pm 2.2\%$ and $82.6 \pm 1.2\%$, respectively. On NICO++, our method also performs best, achieving $87.4 \pm 3.8\%$ worst-group accuracy and $94.0 \pm 0.5\%$ average accuracy.

These results suggest that generative intervention is most effective when paired with a representation-level invariance objective. By aligning variants of the same causal object across different backgrounds, Cross-Variant SSL reduces reliance on spurious context and improves generalization under distribution shift.

\subsection{Ablation Study}
To understand the contribution of each architectural component, we conduct an ablation study on the Waterbirds dataset. We systematically remove components from our full framework and report the resulting worst-group accuracy (extended ablation results are in the Appendix~\ref{sec:extended_ablation} ).

\textbf{Impact of Cross-Variant SSL.}
Bypassing the Stage 1 SSL pretraining and optimizing the ResNet-50 directly on the generated variants using GroupDRO (``w/o Cross-Variant SSL'') results in a substantial drop in worst-group accuracy (from $92.5\%$ to $89.6\%$). This confirms our hypothesis that simply adding generative data is insufficient; the model must be explicitly constrained via contrastive positive pairs to ignore the randomized backgrounds.

\textbf{Impact of ERM Warmup.}
GroupDRO is highly sensitive to early gradient updates. When we disable the 2-epoch ERM warmup (setting $W=0$), the model overfits to noisy, hard-to-learn minority samples before the classifier head has mapped the baseline features. This leads to performance degradation of roughly $4.8\%$.
\section{Conclusion}
\label{sec:conclusion}
We presented a two-stage framework for mitigating spurious correlations by combining generative environmental randomization with representation-level invariance learning. First, a zero-shot segmentation and diffusion inpainting pipeline isolates the foreground object and generates context-shifted variants with diverse backgrounds. Second, Cross-Variant SSL encourages object-centric representations, followed by ERM-warmed GroupDRO fine-tuning with layer-wise learning rates to improve worst-group robustness.

Across Waterbirds, MetaShift, and NICO++, our method achieves strong worst-group accuracies of $92.5\%$, $81.7\%$, and $87.4\%$, respectively. It also obtains the best average accuracies of $95.4\%$, $82.6\%$, and $94.0\%$ on the same benchmarks. These results show that generated variants are more effective when used to learn invariant representations rather than only as additional supervised samples.

A limitation of our approach is its sensitivity to the quality and diversity of generated variants, reflected in variance across seeds. Future work will focus on more stable variant selection and generation-quality filtering.
\section*{Acknowledgment}
The authors acknowledge GPU compute support from the Infosys Center for Artificial Intelligence at IIIT-Delhi.
\section*{LLM Usage}
The authors used an LLM to assist with grammatical and stylistic editing only. All changes were reviewed by the authors, who take full responsibility for the final manuscript.
{
    \small
    \bibliographystyle{ieeenat_fullname}
    \bibliography{main}

@misc{arjovsky2020invariantriskminimization,
      title={Invariant Risk Minimization}, 
      author={Martin Arjovsky and Léon Bottou and Ishaan Gulrajani and David Lopez-Paz},
      year={2020},
      eprint={1907.02893},
      archivePrefix={arXiv},
      primaryClass={stat.ML},
      url={https://arxiv.org/abs/1907.02893}, 
}

@misc{hendrycks2021naturaladversarialexamples,
      title={Natural Adversarial Examples}, 
      author={Dan Hendrycks and Kevin Zhao and Steven Basart and Jacob Steinhardt and Dawn Song},
      year={2021},
      eprint={1907.07174},
      archivePrefix={arXiv},
      primaryClass={cs.LG},
      url={https://arxiv.org/abs/1907.07174}, 
}

@misc{geirhos2022imagenettrainedcnnsbiasedtexture,
      title={ImageNet-trained CNNs are biased towards texture; increasing shape bias improves accuracy and robustness}, 
      author={Robert Geirhos and Patricia Rubisch and Claudio Michaelis and Matthias Bethge and Felix A. Wichmann and Wieland Brendel},
      year={2022},
      eprint={1811.12231},
      archivePrefix={arXiv},
      primaryClass={cs.CV},
      url={https://arxiv.org/abs/1811.12231}, 
}

@misc{beery2018recognitionterraincognita,
      title={Recognition in Terra Incognita}, 
      author={Sara Beery and Grant van Horn and Pietro Perona},
      year={2018},
      eprint={1807.04975},
      archivePrefix={arXiv},
      primaryClass={cs.CV},
      url={https://arxiv.org/abs/1807.04975}, 
}

@misc{sagawa2020distributionallyrobustneuralnetworks,
      title={Distributionally Robust Neural Networks for Group Shifts: On the Importance of Regularization for Worst-Case Generalization}, 
      author={Shiori Sagawa and Pang Wei Koh and Tatsunori B. Hashimoto and Percy Liang},
      year={2020},
      eprint={1911.08731},
      archivePrefix={arXiv},
      primaryClass={cs.LG},
      url={https://arxiv.org/abs/1911.08731}, 
}

@misc{liang2022metashiftdatasetdatasetsevaluating,
      title={MetaShift: A Dataset of Datasets for Evaluating Contextual Distribution Shifts and Training Conflicts}, 
      author={Weixin Liang and James Zou},
      year={2022},
      eprint={2202.06523},
      archivePrefix={arXiv},
      primaryClass={cs.LG},
      url={https://arxiv.org/abs/2202.06523}, 
}

@misc{zhang2022nicobetterbenchmarkingdomain,
      title={NICO++: Towards Better Benchmarking for Domain Generalization}, 
      author={Xingxuan Zhang and Yue He and Renzhe Xu and Han Yu and Zheyan Shen and Peng Cui},
      year={2022},
      eprint={2204.08040},
      archivePrefix={arXiv},
      primaryClass={cs.CV},
      url={https://arxiv.org/abs/2204.08040}, 
}

@misc{liu2024groundingdinomarryingdino,
      title={Grounding DINO: Marrying DINO with Grounded Pre-Training for Open-Set Object Detection}, 
      author={Shilong Liu and Zhaoyang Zeng and Tianhe Ren and Feng Li and Hao Zhang and Jie Yang and Qing Jiang and Chunyuan Li and Jianwei Yang and Hang Su and Jun Zhu and Lei Zhang},
      year={2024},
      eprint={2303.05499},
      archivePrefix={arXiv},
      primaryClass={cs.CV},
      url={https://arxiv.org/abs/2303.05499}, 
}

@misc{kirillov2023segment,
      title={Segment Anything}, 
      author={Alexander Kirillov and Eric Mintun and Nikhila Ravi and Hanzi Mao and Chloe Rolland and Laura Gustafson and Tete Xiao and Spencer Whitehead and Alexander C. Berg and Wan-Yen Lo and Piotr Dollár and Ross Girshick},
      year={2023},
      eprint={2304.02643},
      archivePrefix={arXiv},
      primaryClass={cs.CV},
      url={https://arxiv.org/abs/2304.02643}, 
}

@misc{chen2020simpleframeworkcontrastivelearning,
      title={A Simple Framework for Contrastive Learning of Visual Representations}, 
      author={Ting Chen and Simon Kornblith and Mohammad Norouzi and Geoffrey Hinton},
      year={2020},
      eprint={2002.05709},
      archivePrefix={arXiv},
      primaryClass={cs.LG},
      url={https://arxiv.org/abs/2002.05709}, 
}

@misc{liu2021justtraintwiceimproving,
      title={Just Train Twice: Improving Group Robustness without Training Group Information}, 
      author={Evan Zheran Liu and Behzad Haghgoo and Annie S. Chen and Aditi Raghunathan and Pang Wei Koh and Shiori Sagawa and Percy Liang and Chelsea Finn},
      year={2021},
      eprint={2107.09044},
      archivePrefix={arXiv},
      primaryClass={cs.LG},
      url={https://arxiv.org/abs/2107.09044}, 
}

@misc{kirichenko2023layerretrainingsufficientrobustness,
      title={Last Layer Re-Training is Sufficient for Robustness to Spurious Correlations}, 
      author={Polina Kirichenko and Pavel Izmailov and Andrew Gordon Wilson},
      year={2023},
      eprint={2204.02937},
      archivePrefix={arXiv},
      primaryClass={cs.LG},
      url={https://arxiv.org/abs/2204.02937}, 
}

@misc{yao2022improvingoutofdistributionrobustnessselective,
      title={Improving Out-of-Distribution Robustness via Selective Augmentation}, 
      author={Huaxiu Yao and Yu Wang and Sai Li and Linjun Zhang and Weixin Liang and James Zou and Chelsea Finn},
      year={2022},
      eprint={2201.00299},
      archivePrefix={arXiv},
      primaryClass={cs.LG},
      url={https://arxiv.org/abs/2201.00299}, 
}

@misc{noohdani2024decomposeandcomposecompositionalapproachmitigating,
      title={Decompose-and-Compose: A Compositional Approach to Mitigating Spurious Correlation}, 
      author={Fahimeh Hosseini Noohdani and Parsa Hosseini and Aryan Yazdan Parast and Hamidreza Yaghoubi Araghi and Mahdieh Soleymani Baghshah},
      year={2024},
      eprint={2402.18919},
      archivePrefix={arXiv},
      primaryClass={cs.CV},
      url={https://arxiv.org/abs/2402.18919}, 
}

@misc{parast2025ddbdiffusiondrivenbalancing,
      title={DDB: Diffusion Driven Balancing to Address Spurious Correlations}, 
      author={Aryan Yazdan Parast and Basim Azam and Naveed Akhtar},
      year={2025},
      eprint={2503.17226},
      archivePrefix={arXiv},
      primaryClass={cs.CV},
      url={https://arxiv.org/abs/2503.17226}, 
}

@misc{labs2025flux1kontextflowmatching,
      title={FLUX.1 Kontext: Flow Matching for In-Context Image Generation and Editing in Latent Space},
      author={Black Forest Labs and Stephen Batifol and Andreas Blattmann and Frederic Boesel and Saksham Consul and Cyril Diagne and Tim Dockhorn and Jack English and Zion English and Patrick Esser and Sumith Kulal and Kyle Lacey and Yam Levi and Cheng Li and Dominik Lorenz and Jonas Müller and Dustin Podell and Robin Rombach and Harry Saini and Axel Sauer and Luke Smith},
      year={2025},
      eprint={2506.15742},
      archivePrefix={arXiv},
      primaryClass={cs.GR},
      url={https://arxiv.org/abs/2506.15742},
}

@misc{flux2024,
    author={Black Forest Labs},
    title={FLUX},
    year={2024},
    howpublished={\url{https://github.com/black-forest-labs/flux}},
}

@book{Pearl_2009, place={Cambridge}, edition={2}, title={Causality}, publisher={Cambridge University Press}, author={Pearl, Judea}, year={2009}}

@misc{he2015deepresiduallearningimage,
      title={Deep Residual Learning for Image Recognition}, 
      author={Kaiming He and Xiangyu Zhang and Shaoqing Ren and Jian Sun},
      year={2015},
      eprint={1512.03385},
      archivePrefix={arXiv},
      primaryClass={cs.CV},
      url={https://arxiv.org/abs/1512.03385}, 
}
}

\clearpage
\setcounter{page}{1}
\maketitlesupplementary

\appendix

\section{Related Work}
\label{sec:related_work}

\subsection{Group Robustness without Data Augmentation}

ERM models are known to exploit spurious correlations,
achieving high average accuracy while failing on minority
groups~\cite{sagawa2020distributionallyrobustneuralnetworks,
arjovsky2020invariantriskminimization}. GroupDRO~\cite{
sagawa2020distributionallyrobustneuralnetworks} directly
minimizes the worst-group loss but requires group labels
during training. JTT~\cite{liu2021justtraintwiceimproving}
avoids this by upweighting misclassified samples from an
initial ERM model. DFR~\cite{kirichenko2023layerretrainingsufficientrobustness}
demonstrates that retraining only the final classification
layer on a balanced validation set is sufficient for group
robustness. While effective, these methods introduce no
new samples and struggle when minority groups are
extremely sparse.

\subsection{Data Augmentation for Group Robustness}

To address minority group scarcity, augmentation-based
methods construct new training samples. LISA~\cite{yao2022improvingoutofdistributionrobustnessselective}
interpolates samples and labels across domains. DaC~\cite{
noohdani2024decomposeandcomposecompositionalapproachmitigating}
disentangles causal and non-causal image components via
ERM attribution scores and recombines them to synthesize
minority samples, but is constrained to existing image
components, limiting semantic diversity. DDB~\cite{
parast2025ddbdiffusiondrivenbalancing} extends this with
textual inversion and diffusion-based inpainting to modify
causal object regions, achieving improved semantic control.
However, both methods operate on the causal object and
preserve the original background, which can retain
dataset-specific spurious cues. Our approach instead
targets the background directly, synthesizing entirely
novel environments via structural inpainting to eliminate
background bias at its source.

\subsection{Contrastive Learning for Invariant
Representations}

Contrastive learning frameworks such as SimCLR~\cite{
chen2020simpleframeworkcontrastivelearning} learn
invariant representations by aligning positive pairs
constructed from stochastic augmentations of the same
image. While effective for low-level perturbations,
standard augmentations such as cropping and color jitter
cannot suppress semantic background correlations, as both
views still share the same background context. Our
Cross-Variant SSL addresses this directly by constructing
positive pairs from diffusion-generated variants of the
same causal object placed in entirely different synthesized
backgrounds, explicitly forcing the encoder to discard
background-specific features.

\subsection{Robust Fine-Tuning}

Recent work shows that fine-tuning strategy critically
affects OOD robustness. LP-FT~\cite{kirichenko2023layerretrainingsufficientrobustness}
demonstrates that linear probing before full fine-tuning
preserves pretrained representations and improves
generalization. Motivated by this, our Stage~2 employs
layer-wise learning rates to protect the background-
invariant features learned in Stage~1, combined with an
ERM warmup before activating GroupDRO to prevent
adversarial group weights from amplifying noisy gradients
early in training.

\section{Diffusion Generation Prompts}
\label{sec:generation_prompts}

For each dataset, we condition the FLUX.1-Fill \cite{labs2025flux1kontextflowmatching, flux2024} inpainting model on a fixed set of textual prompts to synthesize diverse, photorealistic background environments. For each training image, the background region (identified by the inverted causal object mask) is independently inpainted once per prompt, yielding four context-shifted variants per sample. The prompts are designed to cover the full range of spurious environments present in each benchmark, as well as novel contexts unseen during training.

\subsection{Waterbirds}

For Waterbirds, the prompts target two semantically distinct background categories: \textit{water} backgrounds and \textit{land} backgrounds , directly targeting the spurious correlation in the dataset.

\begin{table}[h]
\centering
\caption{Generation prompts for \textbf{Waterbirds}.}
\label{tab:prompts_waterbirds}
\setlength{\tabcolsep}{4pt}
\begin{tabularx}{\columnwidth}{cX}
\toprule
\textbf{Cat.} & \textbf{Prompt} \\
\midrule
Water &
\textit{``A highly detailed, photorealistic wide shot of a calm blue ocean with gentle rolling waves, bright sunny sky, natural lighting, 4k''} \\[4pt]
Water &
\textit{``A serene, misty freshwater lake at sunrise, calm water reflecting the golden hour light, tranquil nature photography''} \\[4pt]
Land &
\textit{``A rugged mountain landscape with rocky peaks, sparse alpine vegetation, and a clear blue sky, crisp daylight, high resolution''} \\[4pt]
Land &
\textit{``A dense, lush green forest floor with dappled sunlight filtering through the trees, mossy environment, cinematic lighting, highly detailed''} \\
\bottomrule
\end{tabularx}
\end{table}

\subsection{MetaShift}

For MetaShift, prompts cover \textit{indoor} contexts and \textit{outdoor} contexts. Each majority-group image is inpainted with prompts from the \textit{opposite} class context to generate cross-context minority samples.

\begin{table}[h]
\centering
\caption{Generation prompts for \textbf{MetaShift}.}
\label{tab:prompts_metashift}
\setlength{\tabcolsep}{4pt}
\begin{tabularx}{\columnwidth}{cX}
\toprule
\textbf{Cat.} & \textbf{Prompt} \\
\midrule
Indoor &
\textit{``A highly detailed, photorealistic cozy bedroom interior with a soft bed, natural indoor lighting, realistic home scene, 4k''} \\[4pt]

Indoor &
\textit{``A modern living room with a comfortable sofa, warm daylight, realistic interior photography, highly detailed''} \\[4pt]

Outdoor &
\textit{``A realistic outdoor park scene with a wooden bench, natural daylight, green surroundings, photorealistic, high resolution''} \\[4pt]

Outdoor &
\textit{``A realistic urban outdoor setting featuring a bicycle nearby, natural daylight, photorealistic street scene, highly detailed''} \\
\bottomrule
\end{tabularx}
\end{table}

\subsection{NICO++}

For our NICO++ subset, prompts correspond to the four spurious training contexts (\textit{grass}, \textit{outdoor}, \textit{rock}, \textit{water}). The two OOD test contexts (\textit{autumn} and \textit{dim}) are deliberately excluded from the prompt set to prevent any leakage of test-time distribution information into training.

\begin{table}[h]
\centering
\caption{Generation prompts for \textbf{NICO++}.}
\label{tab:prompts_nicopp}
\setlength{\tabcolsep}{4pt}
\begin{tabularx}{\columnwidth}{cX}
\toprule
\textbf{Context} & \textbf{Prompt} \\
\midrule
Grass &
\textit{``A dense grassy field, endless bright green blades of grass, vibrant lush meadow vegetation, 4k''} \\[4pt]
Outdoor &
\textit{``An empty asphalt highway, cracked gray pavement, urban road landscape with painted lane lines, highly detailed photography, 4k''} \\[4pt]
Rock &
\textit{``A solid rock formation, massive smooth gray boulders and harsh stone surfaces, earthy tones, 4k''} \\[4pt]
Water &
\textit{``Clear water filling a deep water pond, gentle water ripples moving across the water surface. Highly detailed water photography, 4k''} \\
\bottomrule
\end{tabularx}
\end{table}

The deliberate exclusion of \textit{autumn} and \textit{dim} prompts from the NICO++ generation set is a key design choice: by never exposing the generative pipeline to the test contexts, any improvement in worst-group accuracy on the OOD test set reflects genuine background invariance learned by the encoder, rather than memorization of test-time appearances.

\section{Extended Dataset Details}
\label{sec:dataset_details}

We evaluate our framework on three widely used distribution
shift benchmarks:

\begin{itemize}
    \item \textbf{Waterbirds}~\cite{sagawa2020distributionallyrobustneuralnetworks}:
    This dataset combines bird photographs with backgrounds
    from the Places dataset. The training set is spuriously
    correlated such that waterbirds frequently appear on water
    backgrounds and landbirds on land. The test set evaluates
    generalization to the minority groups (waterbirds on land,
    landbirds on water). We adopt the standard splits and
    evaluation protocol identical to our
    baseline~\cite{parast2025ddbdiffusiondrivenbalancing}.

    \item \textbf{MetaShift}~\cite{liang2022metashiftdatasetdatasetsevaluating}:
    We utilize a targeted subset where the training distribution
    establishes a spurious correlation between the ``dog'' class
    and outdoor contexts (benches, bikes), and the ``cat'' class
    with indoor contexts (beds, sofas). The OOD test set
    evaluates both classes within a completely novel context
    (``shelf''). We adopt the standard splits and evaluation
    protocol identical to our baseline~\cite{parast2025ddbdiffusiondrivenbalancing}.

    \item \textbf{NICO++}~\cite{zhang2022nicobetterbenchmarkingdomain}:
    A comprehensive benchmark designed for OOD generalization
    in image classification. We construct a targeted binary
    classification subset using the \textit{fox} and
    \textit{wolf} classes. In the training set, both fox and
    wolf images appear across four spurious background contexts:
    \textit{grass}, \textit{outdoor}, \textit{rock}, and
    \textit{water}. The OOD evaluation split uses entirely
    novel contexts, \textit{autumn} and \textit{dim}, which
    are absent from training. These OOD images are partitioned
    into validation (15\%) and test (85\%) sets via random
    splitting, following the same protocol as MetaShift.
    Worst-group accuracy is computed as the lowest per
    class--background combination accuracy, following the
    same evaluation protocol as MetaShift.
\end{itemize}

\subsection{Dataset Split Statistics}

Tables~\ref{tab:waterbirds_splits}--\ref{tab:nicopp_splits}
report the original sample counts per group across
training, validation, and test splits for all three
benchmarks.

\begin{table}[h]
\centering
\caption{Original group sample counts for \textbf{Waterbirds}.}
\label{tab:waterbirds_splits}
\setlength{\tabcolsep}{3pt}
\begin{tabularx}{\columnwidth}{Xrrr}
\toprule
\textbf{Group} & \textbf{Train} & \textbf{Val} & \textbf{Test} \\
\midrule
Landbird, land   & 3,498 &  467 & 2,255 \\
Landbird, water  &   184 &  466 & 2,255 \\
Waterbird, land  &    56 &  133 &   642 \\
Waterbird, water & 1,018 &  133 &   642 \\
\midrule
\textbf{Total}   & 4,756 & 1,199 & 5,794 \\
\bottomrule
\end{tabularx}
\end{table}

\begin{table}[h]
\centering
\caption{Original group sample counts for \textbf{MetaShift}.}
\label{tab:metashift_splits}
\setlength{\tabcolsep}{3pt}
\begin{tabularx}{\columnwidth}{Xrrr}
\toprule
\textbf{Group} & \textbf{Train} & \textbf{Val} & \textbf{Test} \\
\midrule
Cat, sofa    & 231 &  0 &   0 \\
Cat, bed     & 380 &  0 &   0 \\
Dog, bench   & 145 &  0 &   0 \\
Dog, bike    & 367 &  0 &   0 \\
Cat, shelf   &   0 & 34 & 201 \\
Dog, shelf   &   0 & 47 & 259 \\
\midrule
\textbf{Total} & 1,123 & 81 & 460 \\
\bottomrule
\end{tabularx}
\end{table}

\begin{table}[h]
\centering
\caption{Original group sample counts for our
\textbf{NICO++} subset (fox and wolf). Training contexts
are \textit{grass}, \textit{outdoor}, \textit{rock}, and
\textit{water}. OOD contexts (\textit{autumn}, \textit{dim})
are split randomly into 15\% validation and 85\% test.}
\label{tab:nicopp_splits}
\setlength{\tabcolsep}{3pt}
\begin{tabularx}{\columnwidth}{Xrrr}
\toprule
\textbf{Group} & \textbf{Train} & \textbf{Val} & \textbf{Test} \\
\midrule
Fox,  grass   & 401 & 0 & 0 \\
Fox,  outdoor & 161 & 0 & 0 \\
Fox,  rock    & 152 & 0 & 0 \\
Fox,  water   & 186 & 0 & 0 \\
Wolf, grass   & 239 & 0 & 0 \\
Wolf, outdoor & 325 & 0 & 0 \\
Wolf, rock    & 265 & 0 & 0 \\
Wolf, water   & 277 & 0 & 0 \\
Fox,  autumn  & 0 & 33 & 184 \\
Fox,  dim     & 0 & 20 & 113 \\
Wolf, autumn  & 0 & 35 & 200 \\
Wolf, dim     & 0 & 27 & 152 \\
\midrule
\textbf{Total} & 2,006 & 115 & 649 \\
\bottomrule
\end{tabularx}
\end{table}

\newpage
\subsection{Diffusion-Generated Sample Counts}

For each training image, the causal object is first
isolated via zero-shot segmentation. Images where the
object is not detected are skipped and excluded from training. For all successfully segmented images, four
context-shifted variants are produced by inpainting the
background region with four distinct prompts. Any generated
sample in which the object is no longer detectable in the
output is subsequently discarded. Generation is applied
only to the training split; validation and test splits
remain unmodified.
Tables~\ref{tab:waterbirds_gen}--\ref{tab:nicopp_gen}
report the original and final training set sizes per class
after generation.

\begin{table}[h]
\centering
\caption{Original + Generated sample counts for \textbf{Waterbirds}.}
\label{tab:waterbirds_gen}
\setlength{\tabcolsep}{3pt}
\begin{tabularx}{\columnwidth}{Xr}
\toprule
\textbf{Class, Background} & \textbf{Original + Generated} \\
\midrule
Landbird, land    & 17490 \\
Landbird, water   & 920 \\
Waterbird, land   & 280 \\
Waterbird, water  & 5090 \\
\midrule
\textbf{Total}    & 23780 \\
\bottomrule
\end{tabularx}
\end{table}

\begin{table}[h]
\centering
\caption{Original + Generated sample counts for \textbf{MetaShift}.}
\label{tab:metashift_gen}
\setlength{\tabcolsep}{3pt}
\begin{tabularx}{\columnwidth}{Xr}
\toprule
\textbf{Class, Context} & \textbf{Original + Generated} \\
\midrule
Cat, sofa   & 1708 \\
Cat, bed    & 751 \\
Dog, bench  & 533 \\
Dog, bike   & 1011 \\
\midrule
\textbf{Total} & 4003 \\
\bottomrule
\end{tabularx}
\end{table}

\begin{table}[h]
\centering
\caption{Original + Generated counts for our \textbf{NICO++}
subset (fox and wolf).}
\label{tab:nicopp_gen}
\setlength{\tabcolsep}{3pt}
\begin{tabularx}{\columnwidth}{Xr}
\toprule
\textbf{Class, Context} & \textbf{Original + Generated} \\
\midrule
Fox, grass   & 1,822 \\
Fox, outdoor & 718 \\
Fox, rock    & 639 \\
Fox, water   & 852 \\
Wolf, grass  & 959 \\
Wolf, outdoor & 1,208 \\
Wolf, rock   & 1,065 \\
Wolf, water  & 1,063 \\
\midrule
\textbf{Total} & 8,326 \\
\bottomrule
\end{tabularx}
\end{table}

\newpage

\section{Extended Implementation Details}
\label{sec:implementation_details}

Our pipeline utilizes Grounding DINO with a bounding box
threshold of 0.25, paired with the Segment Anything Model
(SAM) to generate precise binary masks of the causal
objects. The background environments are inverted and
resynthesized using FLUX.1-Fill-dev. For each image, we condition the diffusion model on a set of diverse textual prompts (e.g., ``in a dense forest,'' ``on a busy urban street,'' ``in a modern bedroom'') to generate four distinct variants at a resolution of $1024 \times 1024$. We use FLUX.1-Fill-dev with an inverted SAM background mask to fill only the background while preserving the object, using 20 inference steps and a classifier-free guidance scale of 30.0. After background generation, we paste the original object back onto the generated image using a shrunk and Gaussian-blurred object mask, ensuring cleaner subject preservation and proper blending with the new background.
Generated variants were automatically filtered using the same Grounding DINO detection pipeline. A variant was retained only if the target object was detected above the detection threshold.
For the feature extractor, we utilize a standard ResNet-50
backbone initialized with ImageNet pretraining. During
Stage~1 (Cross-Variant SSL), the projection head is a
2-layer MLP mapping to a 128-dimensional space, trained
for 10 epochs with an InfoNCE loss at temperature
$\tau = 0.2$. During Stage~2, we optimize using AdamW
for 6 epochs with an ERM warmup for the first 2 epochs
before activating the GroupDRO adversarial weight update
with step size $\eta_q = 0.01$. We apply layer-wise
learning rates scaled from a base of $1 \times 10^{-4}$:
$0.1\times$ for early encoder layers, $0.5\times$ for
the final residual block, and $1.0\times$ for the
classifier head. All results are reported as the mean
and standard deviation over three random initialization
seeds.

All hyperparameters are shared across datasets except the
SSL projector hidden dimension used in the Stage~1 projection
head, as reported in Table~\ref{tab:hyperparams}.
All experiments were conducted on an NVIDIA A100 GPU.

\begin{table}[h]
\centering
\caption{Dataset-specific SSL projector hidden dimensions.
All other hyperparameters are shared across datasets:
Stage~1 uses 10 epochs, $\tau{=}0.2$, and projection
dimension 128; Stage~2 uses 6 epochs, ERM warmup for
2 epochs, and $\eta_q{=}0.01$.}
\label{tab:hyperparams}
\setlength{\tabcolsep}{4pt}
\begin{tabularx}{\columnwidth}{Xccc}
\toprule
\textbf{Hyperparameter} & \textbf{Waterbirds} &
\textbf{MetaShift} & \textbf{NICO++} \\
\midrule
SSL projector hidden dimension & 2048 & 512 & 512 \\
\bottomrule
\end{tabularx}
\end{table}

\section{Extended Ablation Study}
\label{sec:extended_ablation}

Table~\ref{tab:extended_ablation} provides the
comprehensive numerical results of our component-wise
ablation study on the Waterbirds dataset. Removing any
single component from our proposed pipeline results in
a distinct degradation of worst-group accuracy,
validating our tightly coupled two-stage architecture.

Notably, the largest single drop comes from removing
layer-wise learning rates ($92.5\% \to 85.8\%$),
confirming that preserving the background-invariant
features learned in Stage~1 is critical during
fine-tuning. Bypassing Cross-Variant SSL pretraining
entirely ($92.5\% \to 89.6\%$) demonstrates that simply
supplying generative data to GroupDRO without the
contrastive invariance objective is insufficient.
Finally, disabling the ERM warmup ($92.5\% \to 87.7\%$)
highlights the sensitivity of GroupDRO to noisy early
gradient updates before the classifier head has
stabilized.

\begin{table}[h]
\centering
\caption{Component-wise ablation on \textbf{Waterbirds}.
Mean worst-group and average accuracy over 3 random seeds.}
\label{tab:extended_ablation}
\setlength{\tabcolsep}{3pt}
\begin{tabularx}{\columnwidth}{Xrr}
\toprule
\textbf{Configuration} & \textbf{Worst} & \textbf{Avg} \\
\midrule
Full Framework (Ours) & $\mathbf{92.5}$ & $\mathbf{95.4}$ \\
\midrule
w/o Cross-Variant SSL   & 89.6 & 93.9 \\
w/o ERM Warmup          & 87.7 & 93.3 \\
w/o Layer-wise LR       & 85.8 & 92.2 \\
\midrule
Baseline (ERM)          & 70.8 & 91.6 \\
\bottomrule
\end{tabularx}
\end{table}
\newpage
\section{Qualitative Results}
\label{sec:qualitative_results}

\begin{figure*}[t]
  \centering
  \includegraphics[width=\textwidth, keepaspectratio]{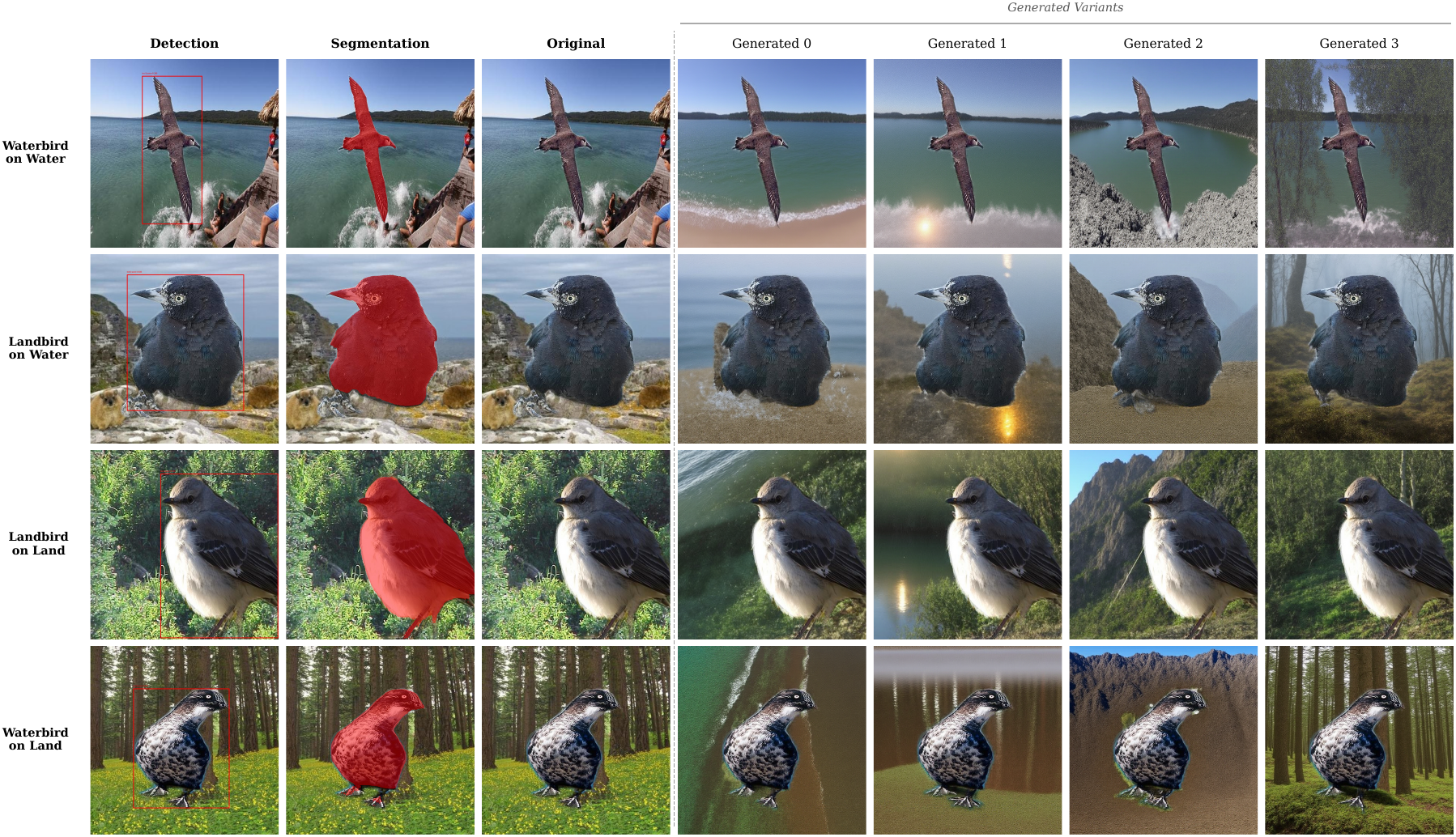}
  \caption{Qualitative results on Waterbirds.}
  \label{fig:qual_waterbirds}
\end{figure*}

\begin{figure*}[t]
  \centering
  \includegraphics[width=\textwidth, keepaspectratio]{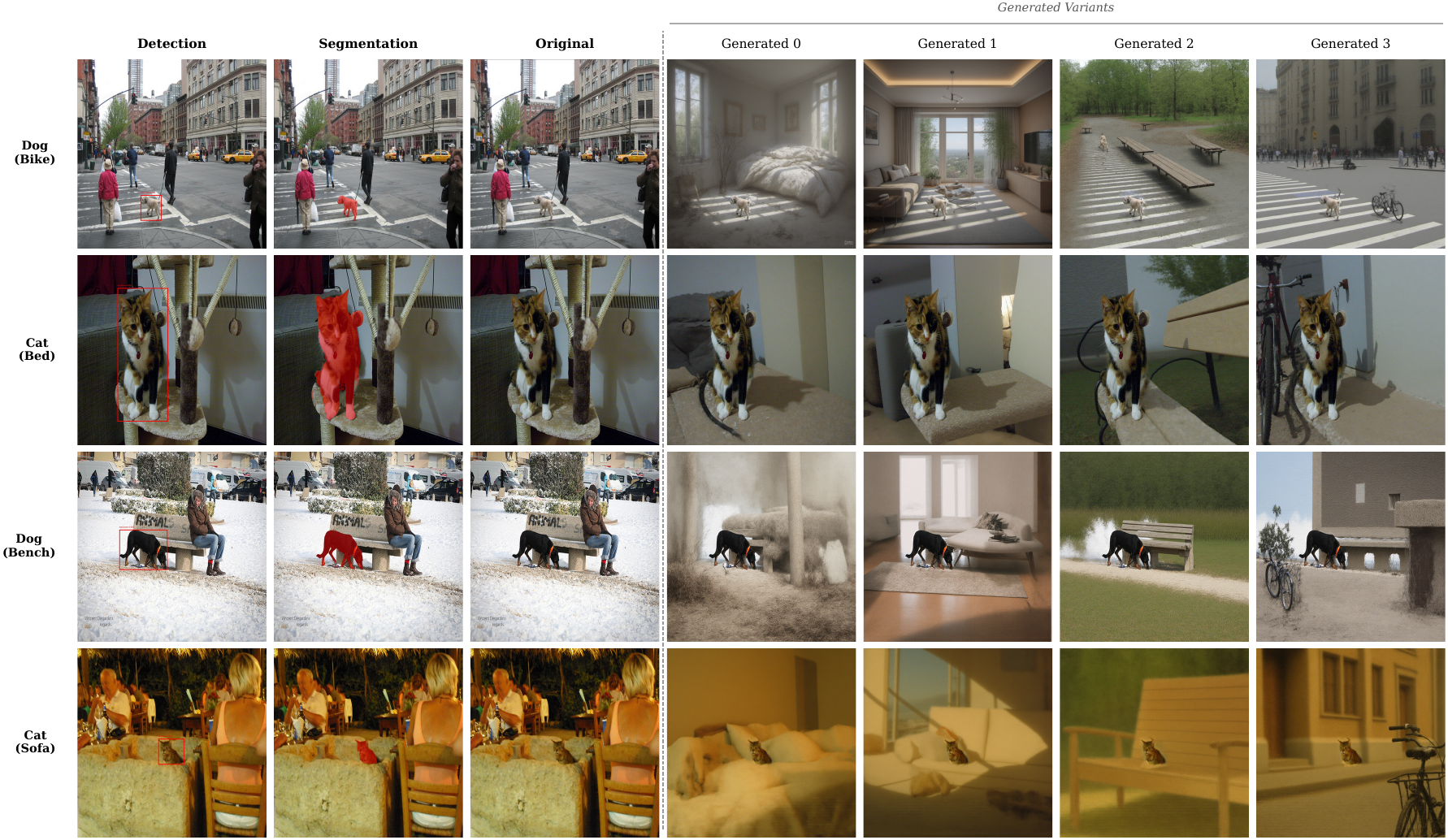}
  \caption{Qualitative results on MetaShift.}
  \label{fig:qual_metashift}
\end{figure*}

\begin{figure*}[t]
  \centering
  \includegraphics[width=\textwidth, keepaspectratio]{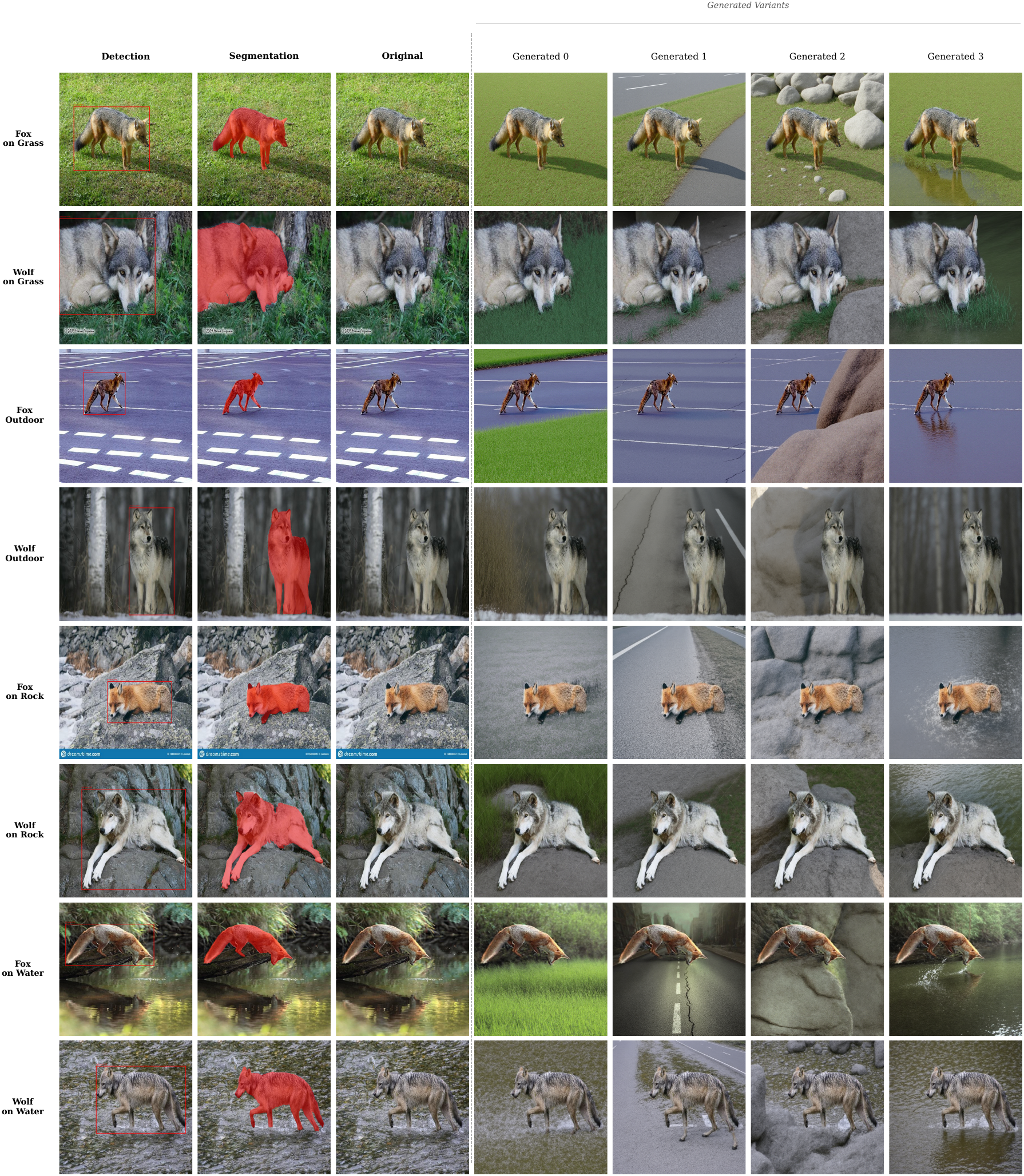}
  \caption{Qualitative results on NICO++.}
  \label{fig:qual_nico}
\end{figure*}

Figures~\ref{fig:qual_waterbirds}--\ref{fig:qual_nico} present qualitative 
examples of our generative pipeline across all three benchmarks. For each 
sample, we show the object detection output, the binary segmentation mask 
produced by SAM, the original image, and four context-shifted variants 
generated by FLUX.1-Fill. Across diverse object categories and background 
types, the foreground subject is consistently preserved while the background 
is fully resynthesized, demonstrating the fidelity and contextual diversity 
of our generation pipeline.

\clearpage

\clearpage

\end{document}